\documentclass[10pt,twocolumn,letterpaper]{article}

\usepackage{cvpr}
\usepackage{times}
\usepackage{epsfig}
\usepackage{graphicx}
\usepackage{amsmath}
\usepackage{amssymb}

\usepackage{enumitem}
\setlist{nosep} 

\usepackage{multirow}
\usepackage{subcaption}
\usepackage{booktabs}
\usepackage[linesnumbered,ruled,vlined]{algorithm2e}

\usepackage{array}
\newcolumntype{H}{>{\setbox0=\hbox\bgroup}c<{\egroup}@{}}

\usepackage[pagebackref=true,breaklinks=true,letterpaper=true,colorlinks,bookmarks=false]{hyperref}

\cvprfinalcopy

\def\cvprPaperID{2090} 

\begin{document}

\title{Consensus-based Sequence Training for Video Captioning}

\author{Sang Phan \quad Gustav Eje Henter \quad Yusuke Miyao \quad Shin'ichi Satoh\\
National Institute of Informatics, Japan\\
{\tt\small \{plsang,gustav,yusuke,satoh\}@nii.ac.jp}
}

\maketitle

\begin{abstract}
	Captioning models are typically trained using the cross-entropy loss. However, their performance is evaluated on other metrics designed to better correlate with human assessments. Recently, it has been shown that reinforcement learning (RL) can directly optimize these metrics in tasks such as captioning. However, this is computationally costly and requires specifying a baseline reward at each step to make training converge. We propose a fast approach to optimize one's objective of interest through the REINFORCE algorithm. First we show that, by replacing model samples with ground-truth sentences, RL training can be seen as a form of weighted cross-entropy loss, giving a fast, RL-based pre-training algorithm. Second, we propose to use the consensus among ground-truth captions of the same video as the baseline reward. This can be computed very efficiently. We call the complete proposal Consensus-based Sequence Training (CST). Applied to the MSRVTT video captioning benchmark, our proposals train significantly faster than comparable methods and establish a new state-of-the-art on the task, improving the CIDEr score from 47.3 to 54.2.
\end{abstract}


\section{Introduction}\label{sect:introduction}

\begin{figure}[t]
	\begin{center}
		\includegraphics[width=1\linewidth]{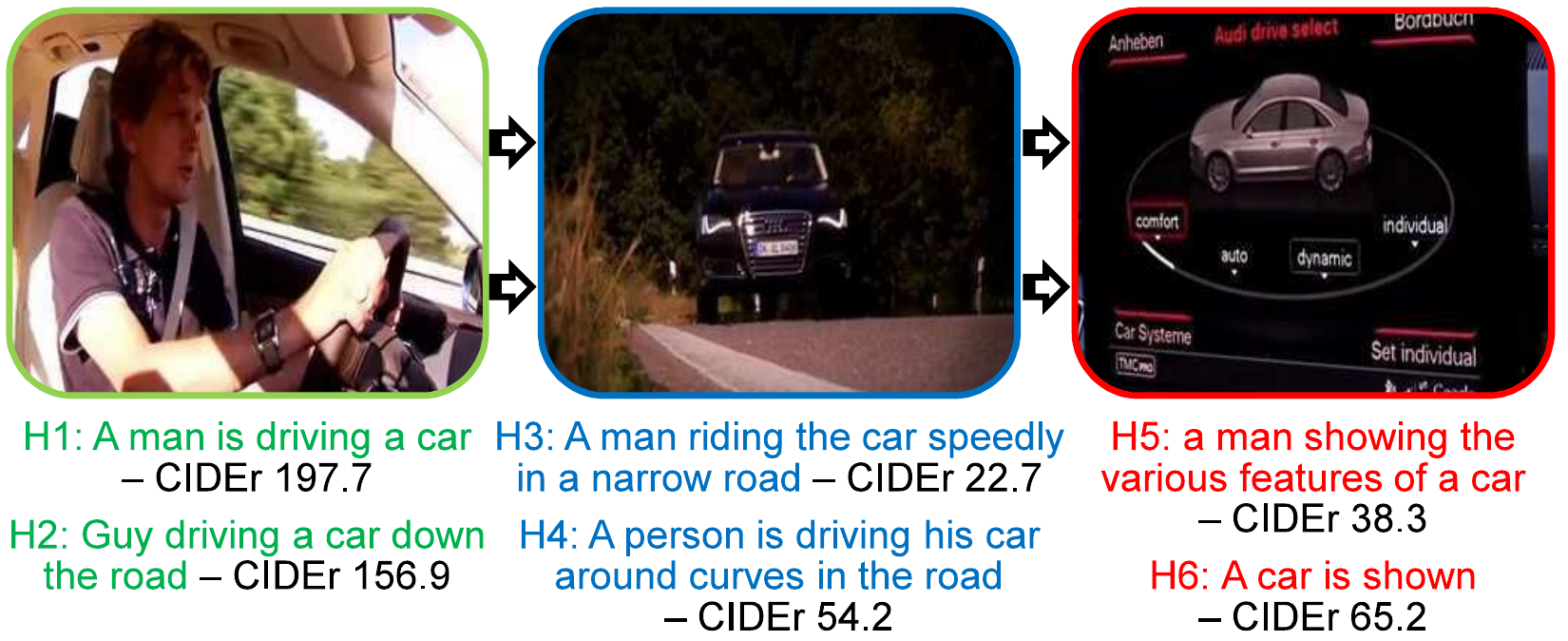}
	\end{center}
	\caption{Examples of human-annotated captions and their different CIDEr consensus scores for a single video.}
	\label{fig_teaser}
\end{figure}

The goal of video captioning is to automatically generate informative and human-like text descriptions of given videos. However, video content is generally very diverse, and so are human-annotated captions. In Fig.~\ref{fig_teaser} we show examples of different descriptions of the same video in the training set of a popular video-captioning dataset. We see that not all captions are created equal -- different persons may pay attention to different segments and aspects of the video, and sometimes annotators may even make mistakes.

Since there is no single ``canonical caption'' for a given video, several works propose automatic evaluation metrics that aim to correlate with human judgment~\cite{papineni2002bleu,lavie2014meteor,lin2004rouge,vedantam2015cider}, enabling the quality of different captions to be compared. Fig.~\ref{fig_teaser} displays the CIDEr score~\cite{vedantam2015cider}, which aims to measure the consensus among annotators. However, there has been less effort to develop captioning systems that directly aim to generate human-like captions by performing well on these metrics.
Instead, most video-captioning systems are trained by the principle of Maximum Likelihood Estimation (MLE), also known as Cross-Entropy (XE) minimization.
This is, unfortunately, not ideal: several works~\cite{basu1998robust,henter2016robust,theis2015note} have shown that MLE/XE emphasizes accurately describing outliers, instead of the most typical case. This makes the outcome of XE training sensitive to unusual or aberrant captions, rather than optimizing for stable output around the human consensus of what an appropriate caption would be.

A compelling alternative to MLE/XE is to maximize the objective of interest directly. This can be done by through the reinforcement learning (RL) framework~\cite{sutton1998reinforcement} using a 
method such as REINFORCE~\cite{williams1992simple}. In RL, the score of a candidate sentence is used as a reward signal, and the model attempts to maximize this reward.
If the chosen reward metric is the CIDEr score~\cite{vedantam2015cider}, which is designed to approximate human judgment of appropriateness and consensus, the model may be taught to generate more human-like captions. 
However, REINFORCE training is difficult because it requires estimating the gradient of the expected reward, which is not stable using mini-batch estimation.

To our knowledge, only one published work, Pasunuru \& Bansal~\cite{pasunuru2017reinforced}, has applied REINFORCE to video captioning. In the process, they obtained top scores on a standard captioning benchmark.
They used a MIXER scheme~\cite{Ranzato2016ICLR}, that gradually mixes RL training into XE training, to stabilize the learning. This is known to be very sensitive to the annealing schedule used~\cite{Liu2016Spider}.

MLE/XE and RL training each have pros and cons. XE training is much faster than RL, but is unlikely to produce human-like captions. RL training can directly optimize objective functions that reward human-likeness, but the gradient with respect to the expected reward is unstable without costly variance reduction. In this paper, we propose a new training approach that combines the advantages of both training schemes. Our main contributions are:
\begin{enumerate}
	\item We show that training with the XE objective is a special case of REINFORCE training when applied to sentences obtained from the ground-truth data, rather than samples from the model being trained.
	\item Based on the previous insight, we introduce a simple weighted XE pre-training scheme (WXE) that approximates RL training. Our proposal is equally stable and fast to train as conventional XE, yet provides a 2-point improvement in terms of the CIDEr metric. Our pre-training addresses a well-known problem of XE known as objective mismatch~\cite{Ranzato2016ICLR}, but (unlike standard RL) not a related issue known as exposure bias (see Sec.~\ref{sect:relatedwork}), thus allowing the losses incurred by these two important problems to be assessed separately.
	\item We further propose to fine-tune our pre-trained models through full REINFORCE, but using the consensus scores of ground-truth captions as the baseline reward. We call this the Self-Consensus Baseline, or SCB. SCB trains twice as fast as estimating the baseline reward through the greedy method~\cite{Rennie2017CVPR}, and provides greater CIDEr score improvements. Together, pre-training and fine-tuning address both objective mismatch and exposure bias, and form our proposed Consensus-based Sequence Training, or CST.
\end{enumerate}
We conduct thorough experiments to demonstrate the advantage of CST on the large-scale MSRVTT~\cite{Xu:CVPR16} benchmark. Our results on the MSRVTT video-captioning dataset improve the results in terms of CIDEr from 47.3 (XE) to 54.2 (full CST), surpassing the best previously published score of 51.7 to establish a new state-of-the-art on the task.

\section{Related Work}\label{sect:relatedwork}

\textbf{Issues with Standard Captioning Models}
In line with other deep text-generation models, captioning models are typically trained to maximize the likelihood of the next word given the previous ground-truth input words (i.e., MLE/XE). This approach has two main drawbacks:

First is the \textit{objective mismatch} problem. This arises because the objective function optimized at training time is not the true objective we are interested in, or rather, is not equal to the metric used to quantify success on the task. Conventional maximum likelihood, in particular, is especially poorly suited for practical applications. Mathematically, it is similar to minimizing a weighted squared error in which accuracy on the least probable captions is given the greatest weight~\cite{basu1998robust}. As highlighted in~\cite{henter2016robust}, this a poor fit for data-generation applications like captioning, since it means that accuracy at the maximum-probability point -- the greedy output eventually returned by the system -- is given the lowest priority of all during training. It also is sensitive to poor or erroneous captions in the data; is is not \textit{statistically robust}~\cite{huber2011robust}.
Randomly sampled (rather than greedy) output is also affected, and models fit using MLE are likely to produce samples that humans perceive as unnatural~\cite{theis2015note}.

To overcome the objective mismatch, one would want to train models to directly optimize the performance metric of interest, which in captioning typically is a discrete NLP metric such as BLEU~\cite{papineni2002bleu}, METEOR~\cite{lavie2014meteor}, ROUGE-L~\cite{lin2004rouge} or CIDEr~\cite{vedantam2015cider}.
Unfortunately, these metrics operate on full sentences (in contrast to model output, which is generated iteratively over words) and are not differentiable, which prevents training using stochastic gradient descent.

The second problem is \textit{exposure bias}~\cite{Ranzato2016ICLR}, which is the input distribution mismatch between training and testing time. During training, the model is only exposed to word sequences from the training data (so called teacher forcing), while at testing time, the model instead only has access to its own predictions; therefore, whenever the model encounters a state it has never been exposed to before, it may behave unpredictably for the rest of the output sequence. Methods such as scheduled sampling~\cite{Bengio2015NIPS} propose to gradually expose models to input words from the model distribution rather than the ground truth.
However, a major limitation of this and related approaches like professor-forcing~\cite{lamb2016professor} is that, at each time-step, the output word needs to be same as in the ground truth, implicitly discouraging the model from generating novel captions.

\textbf{REINFORCE Baseline Estimators} 
Recently, the REINFORCE~\cite{williams1992simple} method has been applied to captioning tasks \cite{Ranzato2016ICLR,Liu2016Spider,hendricks2016generating,Rennie2017CVPR,Anderson2017,pasunuru2017reinforced}. 
This is an RL algorithm that can optimize any metric of interest and trains on sampled sequences, thus overcoming the limitations of maximum likelihood, scheduled sampling, and professor forcing above. Training with REINFORCE is considerably difficult because the gradient of the expected reward is estimated using a single sample, which exhibits high variance.
By estimating an expected baseline reward and subtracting this baseline from the estimated gradient, one can reduce the variance considerably without changing the expected gradient~\cite{williams1992simple}. Most existing work use parametric functions to estimate the expected baseline reward~\cite{Ranzato2016ICLR,Liu2016Spider}; this can be implemented as a separate neural network which takes the hidden state of the captioning model (neural network) as input at each time step.
Instead of estimating the reward signal, Self-Critical Sequence Training (SCST)~\cite{Rennie2017CVPR} utilizes the (greedy) output of the current model at inference time as the baseline, at the cost of having to compute the score of each baseline sequence.
Taking a cue from the up-down model~\cite{Anderson2017}, our proposal trains on multiple captions of the same video simultanelously; however, we propose a Self-Consensus Baseline (SCB), which simply computes the average reward among ground-truth captions as the baseline, rather than generating and scoring a new greedy baseline for every sentence and epoch during training as in~\cite{Rennie2017CVPR}.

\textbf{Consensus-based Caption Evaluation} 
As human evaluation is expensive, captioning performance is commonly assessed using automatic measures that approximate human judgments. We will use the leading CIDEr metric~\cite{vedantam2015cider}, designed to quantify similarity to the consensus among captions better than other metrics. Optimizing CIDEr should encourage output that is similar to typical human captions.


\section{Captioning Models}\label{sect:manet}

\begin{figure*}[t]
	\begin{center}
		\includegraphics[width=1\linewidth]{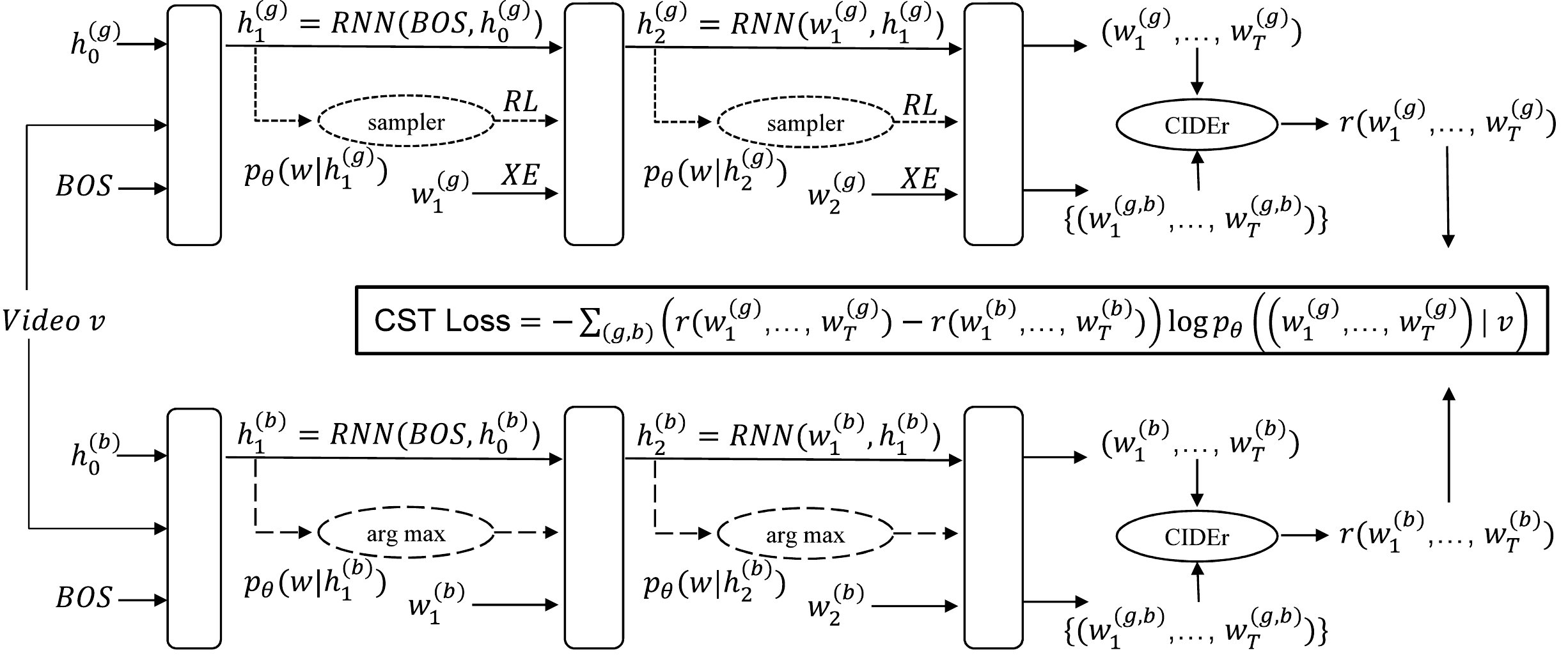}
	\end{center}
	\caption{Overview of Consensus-based Sequence Training (CST). CST can be used to train both XE and RL schemes. Note that the baseline reward is estimated using the consensus scores of \textit{training} captions, e.g., $(g)$ and $(b)$, which is very efficient. 
		Dashed connections indicate the SCST method~\cite{Rennie2017CVPR}.}
	\label{fig:framework}
\end{figure*} 


Most captioning models are based on encoder-decoder neural networks. The encoder takes pre-extracted multimodal features from the video and embed these into a fixed-size space, with the final video representation $v$ being the concatenation of all the embedded features. 
The decoder receives video features from the encoder and generates an output sentence using a Recurrent Neural Network (RNN). 
Among RNNs, the Long Short-Term Memory (LSTM)~\cite{hochreiter1997long} architecture is widely used, since it yields state-of-the-art results in many different applications. At a high level, the computation at each time-step of the LSTM can be expressed in terms of the previously generated word $w_{t-1}$ and hidden state vector $h_{t-1}$ as follows:
\begin{equation}
	h_t = LSTM(h_{t-1}, w_{t-1})
\end{equation}

Let $o_t = W_o h_t$ be the vocabulary-sized output of the LSTM, let $\theta$ denote the parameters of the model, and suppose $h_t$ maintains information on the input video and the previous words, the distribution of the next word is then given by a softmax function:
\begin{equation}
	p_\theta(w_t | h_t) = \text{softmax}(o_t)
\end{equation}


At training time, $w_t$ is typically a ground-truth token; however, one can also perform sampling from the $p_\theta(w_t | h_t)$ distribution. At testing time, $w_t$ is often selected by a greedy, iterative approach such as beam search~\cite{sutskever2014sequence}, which aims to identify the most likely caption (word sequence). 

\subsection{Cross Entropy Training (XE)}
Let $w = (w_1, w_2, ...,w_T)$ be a target, ground-truth sequence. Typically the parameters $\theta$ of the model are learned by minimizing the cross entropy (XE) loss:
\begin{align}
	L^{XE}(\theta) &= - \log p_\theta(w) \\ 
	&= - \sum_{t=1}^{T} \log p_\theta(w_t | h_t)
	\label{eq:xeloss}
\end{align}


\subsection{REINFORCE Training (RL)}
The captioning process can be formulated as a Reinforcement Learning problem~\cite{Ranzato2016ICLR}, where the agent is the LSTM language model interacting with an environment of words and image/video features. The model $p_{\theta}$ defines a policy network, in which each action corresponds to predicting the next word.

Mathematically, the goal of REINFORCE training is to minimize the negative expected reward of model samples:
\begin{equation}
	L^{RL}(\theta) = - \mathbb{E}_{w^s \sim p_{\theta}}[r(w^s)] = - \sum_{w^s} r(w^s)p_\theta(w^s),
	\label{eq:negexpreward}
\end{equation}
where $w^s = (w^s_1, w^s_2, ...,w^s_T)$, with $w^s_t$ being the word sampled from the model $p_\theta$ at time $t$.
To minimize the negative expected reward, one might differentiate \eqref{eq:negexpreward} as:
\begin{equation}
	\nabla_\theta \mathbb{E}_{w^s \sim p_{\theta}}[r(w^s)] = \mathbb{E}_{w^s \sim p_{\theta}}[\nabla_\theta r(w^s)]
\end{equation}
However, this direct approach runs into a problem if the reward function $r(w^s)$ is non-differentiable, i.e., $r(w^s)$ is not a continuous function with respect to $\theta$.
Alternatively, one can differentiate the sum in \eqref{eq:negexpreward} as:
\begin{align}
	\nabla_\theta \mathbb{E}_{w^s \sim p_{\theta}}[r(w^s)] &= 	\nabla_\theta \sum_{w^s} r(w^s)p_\theta(w^s) \\
	&= \sum_{w^s} r(w^s) \nabla_\theta p_\theta(w^s) \\
	&= \sum_{w^s} {p_\theta(w^s)} r(w^s) \frac{\nabla_\theta p_\theta(w^s)}{p_\theta(w^s)} \\
	&= \mathbb{E}_{w^s \sim p_{\theta}}[r(w^s) \nabla_\theta \log p_\theta(w^s)]
	; \label{eq:policygrad}
\end{align}
this expression does not require differentiating $r(w^s)$ w.r.t.\ $\theta$ and can be estimated through Monte Carlo sampling.


In practice, gradients estimated based on \eqref{eq:policygrad} are unstable and it is necessary to perform variance reduction using a baseline estimator $b$ (cf.~\cite{Ranzato2016ICLR}): \begin{align}
	\nabla_\theta L^{RL}(\theta) &= - \mathbb{E}_{w^s \sim p_{\theta}}[(r(w^s) -b) \nabla_\theta \log p_\theta(w^s)]
	\label{eq:addbaseline}\\
	&\approx - (r(w^s) - b) \nabla_\theta \log p_\theta(w^s)
	\label{eq:reinforce}
\end{align}
It is easy to show that the gradient in \eqref{eq:addbaseline} is unchanged as long as $b$ does not depend on the sample $w^s$.

\section{Consensus-based Sequence Training}
This section details the novel contributions of this paper. First, we present a new theoretical connection between XE and RL training. We then expand on this insight to formulate an entire training scheme that utilizes the consensus among the multiple ground-truth captions for each video to improve training in several ways. As a bonus, the two stages of our scheme disentangle the objective mismatch and the exposure bias problems described in~\cite{Ranzato2016ICLR}.
\subsection{Relation between XE Training and RL Training}
\label{sec:rlxe}

Using the chain rule, the derivatives of both the XE and RL loss w.r.t.\ the parameters $\theta$ are:
\begin{gather}
	\frac{\partial L(\theta)}{\partial \theta} = \sum_{t=1}^{T} \frac{\partial L(\theta)}{\partial o_t} \frac{\partial o_t}{\partial \theta} 
\end{gather}
The partial derivative of each loss w.r.t.\ the softmax input $o_t$ is given by~\cite{Ranzato2016ICLR}:
\begin{gather}
	\frac{\partial L^{XE}(\theta)}{\partial o_t} = p_\theta(w_t | h_t) - 1(w_t)
	\label{eq:dxe} \\
	\frac{\partial L^{RL}(\theta)}{\partial o_t} = (r(w^s) - b)(p_\theta(w^s_t | h_t) - 1(w^s_t))
	\label{eq:drl}
\end{gather}
The gradient is the difference between the prediction and the one-hot representation, represented by the $1(\cdot)$ indicator function, of the target word.\footnote{We provide a full derivation in the supplementary material.} In RL training, this gradient is further weighted by a stabilized reward $r(w^s) - b$. In that case the sampled word $w^s_t$ can be either encouraged or penalized, in contrast to the ground-truth word $w_t$ that's always being encouraged in XE training.


There are interesting similarities between Eq.~\eqref{eq:drl} and Eq.~\eqref{eq:dxe}.
In particular, in the situation $w^s_t = w_t$ we obtain:
\begin{align}
	\frac{\partial L^{RL}(\theta)}{\partial o_t} &= (r(w) - b) \frac{\partial L^{XE}(\theta)}{\partial o_t} ,
\end{align}
Therefore, when training RL on the \emph{ground-truth captions}, we can compute the terms in the $L^{RL}(\theta)$ loss as follows:
\begin{align}
	L^{RL}(\theta|w) &= (r(w) - b) L^{XE}(\theta|w),
	\label{eq:rlxe}
\end{align}

Eq.~\eqref{eq:rlxe} suggests that RL training on ground-truth captions (which happens when the captioning system behaves like the human annotators) is a generalization of XE training where each sentence loss is weighted by a reward. More generally, we expect that systems with human-like output can be trained to an approximate optimum on ground-truth captions $w_t$ without having to draw samples $w^s_t$ and evaluate their reward. In cases where sampling or reward calculation is a computational bottleneck (like with CIDEr), this can produce a substantial speed improvement. The trade-off is that no exploration of non-ground-truth sentences is performed, so resulting models may still suffer from exposure bias; however, the weighting still mitigates objective mismatch, meaning that the $w^s_t \approx w_t$ approximation neatly separates the two issues originally recognized in XE, and represents a middle ground between XE and REINFORCE.

Mathematically speaking, how can a weighted cross-entropy be helpful for learning human consensus? Consider a video $v$ described by the caption set $S = \{w^{(1)},..,w^{(N_v)}\}$, $N_v$ being the number of captions. Then:
\begin{equation}
	L^{RL}_v(\theta) = - \frac{1}{N_v}\sum_{i=1}^{N_v} (r(w^{(i)}) - b^{(i)}) \log p_\theta(w^{(i)})
	\label{eq:wxe}
\end{equation}
Intuitively, this loss increases the probability of captions that have high-rewards.
If we choose $r$ to be the CIDEr metric -- designed to measure the consensus among captions -- this loss assigns lower weight to the model's log-likelihood performance on captions with low consensus score (low reward), reducing their impact on the training outcome, in favor of consistently generating more typical captions. This differs from conventional, unweighted XE/MLE, where models typically incur severe log-likelihood penalties for failing to explain bad or atypical captions in training data.

%

\subsection{Self-consensus Baseline}
\label{ssec:scb}
Notably, our pre-training (unlike traditional RL) is stable to train without a baseline $b$, since the use of ground-truth captions significantly constrains the sentence space.
Full RL training, however, still requires a baseline estimator for variance reduction.
Observing that (1) the theoretically optimal baseline $b$ is the expected reward of the current policy and (2) leading automatic captioning systems achieve broadly similar CIDEr scores as the average of held-out human captions (52.9 vs.\ 50.2 in~\cite{Jin2017ACMMultimedia}), we propose using the rewards of ground-truth sentences as the baseline. Specifically, we use the average reward of all the sentences describing the same video to compute $b$. Using the notation in~\ref{sec:rlxe}, the baseline reward for each video $v$ is given by: 
\begin{equation}
	b(v) = \frac{1}{N_v}\sum_{i=1}^{N_v} r(w^{(i)}) 
	\label{eq:scb}
\end{equation}
The cost to estimate our baseline during training is negligible as the rewards can be pre-computed. Since this baseline leverages the mutual consensus among captions of the same video, we call it the Self-Consensus Baseline, or SCB.

\subsection{Full Training Scheme Proposal}
\label{ssec:wxeandscb}
We unite our proposals of an improved pre-training stage and consensus-based SCB in RL training under the label Consensus-based Sequence Training, or CST. Figure~\ref{fig:framework} depicts how the SCB is computed in CST and the main difference with SCST~\cite{Rennie2017CVPR}, while Algorithm~\ref{algo:cst} summarizes the complete CST method.
At first, we propose to ``warm start'' the model by pre-training it on the ground-truth data using the weighted cross-entropy loss defined in Eq.~\eqref{eq:wxe}. This is similar in implementation to standard XE training, except that the log-likelihood of each ground-truth word now is weighted by the CIDEr score of the corresponding caption. We call this method WXE, for Weighted Cross-Entropy. Warm-starting is typical RL practice to prevent models from making random actions in the exponentially large sentence-distribution space at the beginning of training, except that conventional implementations pre-train the model using the unweighted XE loss. As we show in Sec.~\ref{sect:experiments}, our new WXE warm-start strategy gives better objective scores than XE training while being equally fast, and provides a better starting point for later, full REINFORCE training.

After pre-training with WXE, we perform full RL training, in which candidate captions are sampled from the estimated model distribution. This is significantly slower per epoch due to the need to compute the CIDEr scores of model samples, but enables the model to explore more of the possible sentences that can be generated. 

\begin{algorithm}[t]
	\KwData{$\mathbb{D} = \{(v^n,w^n)\}\text{, with }n = 1...N$, is the set of video-caption pairs;}
	Pre-compute the CIDEr scores of all captions in $\mathbb{D}$;\\
	\For{\underline{each epoch}}{
		\For{\underline{each video $v$}}{
			\uIf{\underline{WXE (Pre-)Training}}{
				Get caption $w$ from the ground truth;\\
				Get $r(w)$ from the pre-computed scores;\\
				Set $b=0$;\\
			}
			\ElseIf{\underline{RL Training}}
			{
				Generate sample sequence $w \sim p_{\theta}(.|v)$;\\
				Compute $r(w)$ using CIDEr;\\
				Compute baseline reward $b$ using Eq.~\eqref{eq:scb};\\
			}
			Compute $L^{RL}_v(\theta)$ using Eq.~\eqref{eq:wxe};\\
			Back-propagate to compute $\frac{\partial L^{RL}(\theta)}{\partial \theta}$; \\
			SGD-update $\theta$ using $\frac{\partial L^{RL}(\theta)}{\partial \theta}$;
		}
	}
	\caption{CST}
	\label{algo:cst}
\end{algorithm}

By dividing training into two steps, (1) pre-training using weighted cross-entropy on the ground truth; and (2) RL training on samples from the model, we address both issues of conventional XE training. In (1), we handle the objective mismatch by directly optimizing the CIDEr metric, though still on the ground-truth data. The resulting performance illustrates how much we may improve the captioning model by simply changing the training objective from XE towards CIDEr, without affecting training time. In (2), we allow the model to generate samples and provide feedback through the reward signal. This additionally addresses the exposure bias of MLE, since the sequences evaluated now are sampled from the captioning model.


\section{Experimental Setup}\label{sect:experimental_setup}

\subsection{Dataset and Features}
\label{ssec:datafeats}
We evaluated the performance of our CST method and its variants on the Microsoft Video-to-Text dataset (MSRVTT)~\cite{Xu:CVPR16} popular in captioning. This dataset is composed of 10,000 videos in 20 different categories. Each video is annotated by 20 different persons. We used the data splits provided in~\cite{Xu:CVPR16} which contain 6,513, 497, and 2,990 videos for training, validation, and testing, respectively.

The following features from different modalities were extracted from the videos and used as system inputs:
\begin{itemize}
	\item ResNet~\cite{he2015deep} is an extremely deep network to extract static image features. We used the ResNet 200-layer model trained on the ImageNet dataset to extract image feature from the fc7 layer (2,048-dim) every 8 frames.
	\item C3D~\cite{tran2015learning} captures short-term motion over 16 video frames. We extracted features at the fc6 layer of a model trained on the Sports-1M dataset, and applied mean pooling to represent each video chunk.
	\item MFCC~\cite{davis1980comparison} acoustic features were extracted from 25 ms audio segments with a frame step of 10 ms. The 13 first MFCC coefficients and their $\Delta$ and $\Delta^2$ features were then encoded with a codebook size of 256 clusters using VLAD~\cite{jegou2012aggregating}, yielding a 9,984-dim representation.
	\item Category embedding. Each video in the MSRVTT dataset belongs to one of 20 categories such as music, food, or gaming~\cite{Xu:CVPR16}. We used the word-embedding model proposed in GloVe~\cite{pennington2014glove} to encode the category label of each video into a 300-dimensional vector.
\end{itemize}

\subsection{Implementation Details}
\label{ssec:metrics}
A vocabulary of 10,533 words was created from all words appearing more than thrice in the data. Three special tokens were added to the vocabulary: $<$BOS$>$, $<$EOS$>$, and $<$UNK$>$, representing the beginning and end of a sentence, and unknown tokens, respectively. We did not apply any text pre-processing except for lowercase conversion. Word vectors of size 512 were used to encode previously-generated words (actions) for input to the LSTMs when generating the next word. We use a single-layer unidirectional LSTM with a 512-dimensional hidden state for captioning. Following~\cite{zaremba2014recurrent}, we inserted a rate-0.5 dropout layer on the input and output of each LSTM unit during training.

In each iteration, the encoder loaded a mini-batch of 64 videos with pre-extracted multimodal features. Each feature was then projected into an embedding using a linear layer of dimension 512. We trained on all 20 ground-truth/sampled captions of each video at the same time. The Adam~\cite{kingma2014adam} optimizer was used for training, with a fixed learning rate of $1\times10^{-4}$ in all experiments. XE/WXE pre-training was ran for maximum 50 epochs; full RL training continued until epoch 200, with the best result reported.

We used beam search~\cite{sutskever2014sequence} with a beam size of 5 to find highly probable output sentences for evaluation.
Training and evaluation were implemented with PyTorch, with objective scores calculated using the MS COCO toolkit~\cite{chen2015microsoft}.\footnote{https://github.com/tylin/coco-caption}

\subsection{Training Methods Considered}
\label{ssec:systems}
For the experiments, we trained the same basic captioning LSTM using a number of variants of CST, gradually stepping from conventional XE/MLE training to full CST. 
The different training configurations are introduced below, with an overview provided in Table~\ref{tab:overview}:

\begin{table*}[t]
	\centering
	\caption{Overview of the most important training methods considered in the experiments.}
	\label{tab:overview}
	\begin{tabular}{@{}l|l|l|Hl|l@{}}
		\toprule
		Label & Start from & Sequences used in training & Weighted? & Baseline $b$ & Comment \\
		\midrule
		XE & Scratch & Ground truth & No (not RL) & None & Unweighted log-likelihood; bottom line \\
		CST\_GT\_None & Scratch & Ground truth (``GT'') & Yes & None & Weighted log-likelihood; a.k.a.\ ``WXE'' \\
		CST\_MS\_Greedy & WXE & Model samples (``MS'') & Yes & Greedy~\cite{Rennie2017CVPR} & \\
		CST\_MS\_SCB & WXE & Model samples & Yes & SCB & ``Full CST'' \\
		\midrule
		CIDEnt-RL~\cite{pasunuru2017reinforced} & XE & Truth \& samples (MIXER) & Yes & Lin.\ reg.\ ~\cite{Ranzato2016ICLR} & Previous best \\
		SCST~\cite{Rennie2017CVPR} & XE & Model samples & Yes & Greedy & Our implementation\\
		\bottomrule
	\end{tabular}
\end{table*}

\textbf{XE}: This LSTM was trained on our extracted features and network setup using conventional maximum-likelihood, as a comparison point, and a bottom line against which to judge our improvements.

\textbf{CST\_GT\_None}: This is the CST pre-training paradigm, in which the reward-weighted cross entropy in Eq.~\eqref{eq:wxe} is used for training; for this reason, this training method is also referred to as \textbf{WXE}, for short. This weighting alleviates the objective mismatch problem of XE. The difference from full REINFORCE is that the loss is evaluated on ground-truth captions (``GT'') rather than captions sampled from the model being trained.
(The word ``None'' signifies that no baseline $b$ was used.)



\textbf{CST\_MS\_SCB}: This is a full REINFORCE (RL) training paradigm, where training gradients are computed on model samples (``MS'') as in Eq.~\eqref{eq:reinforce}. Using samples instead of ground-truth captions should address both the objective mismatch and the exposure bias problems simultaneously. For faster convergence, CST\_MS\_SCB training was always initialized from CST\_GT\_None/WXE above. Since REINFORCE training is unstable out of the box in video captioning, we used a baseline $b$ based on self-consensus (SCB introduced in Sec.~\ref{ssec:scb}) -- the average held-out CIDEr score of ground-truth captions -- to stabilize training. This is significantly simpler than the MIXER approach~\cite{Ranzato2016ICLR} used in~\cite{pasunuru2017reinforced}, needing no mixing scheme tuning or annealing.


\textbf{CST\_MS\_Greedy}: This is another full RL training paradigm, identical to CST\_MS\_SCB except that instead of basing $b$ on the consensus among ground-truth captions, $b$ was set equal to the score (reward) of a highly-likely sentence generated by iteratively (greedily) generating the most probable word under the model, as proposed in~\cite{Rennie2017CVPR}.

\textbf{SCST}: This is the same as CST\_MS\_Greedy, except that (like in~\cite{Rennie2017CVPR}) full RL training starts from conventional XE instead of WXE. It thus constitutes a reimplementation of~\cite{Rennie2017CVPR}, but applied to video rather than image captioning.



\section{Experimental Results}\label{sect:experiments}

\subsection{Performance on Different Feature Sets}
To begin with, we performed a preliminary experiment comparing conventional XE and CIDEr-weighted WXE when trained on each of the features in Sec.~\ref{ssec:datafeats} separately, as well as on the full, multimodal feature set. The resulting systems were compared in terms of seven different objective metrics (four BLEU scores and three other metrics, including CIDEr), with the numerical results tabulated in supplementary material.
Except when training only on the Category features, WXE-models outperformed XE baselines in all aspects, confirming the efficacy of the WXE pre-training scheme over conventional XE.
Since the full feature set gave the best performance across the board, we used this input feature set in all subsequent experiments.


\begin{table}[t]
	\centering
	\caption{Per-batch training time for different methods.}
	\label{table_time}
	\begin{tabular}{@{}l|r@{}}
		\toprule
		Model & Wall-clock time (s)\\ \midrule
		XE & \hphantom{0}0.677          \\
		CST\_GT\_None & \hphantom{0}0.594          \\
		CST\_MS\_Greedy & 10.377         \\
		CST\_MS\_SCB & \hphantom{0}5.256          \\ \bottomrule
	\end{tabular}
\end{table}

\subsection{Training Time}
Having selected the feature set, we trained systems using each of the methods in Sec.~\ref{ssec:systems}. Table~\ref{table_time} reports the wall-clock time per batch for training each of these systems except SCST, which performed similarly to CST\_MS\_Greedy. The major computational bottleneck is evaluating CIDEr scores. The WXE and XE methods are by far the fastest (and essentially equally fast), since any required CIDEr scores can be pre-computed. The computation of the reward (CIDEr score) of sentences sampled in the inner loop of CST\_MS makes these full RL schemes run much slower. The fact that the Greedy baseline of~\cite{Rennie2017CVPR} also must compute the CIDEr score of the generated baseline sentence essentially doubles the computational load of CST\_MS\_Greedy and SCST over our proposed CST\_MS\_SCB.



\begin{table}[t]
	\centering
	\caption{Comparison between CST variants and leading previous results achieved on the MSRVTT
		test set.}
	\label{tab:sota}
	\setlength\tabcolsep{3.3pt}
	\begin{tabular}{@{}l|cccc@{}}
		\toprule		
		Method             & \small{BLEU\_4} & \small{METEOR} &
		\small{ROUGE\_L} & \small{CIDEr} \\ \midrule
		\multicolumn{5}{c}{XE and WXE training} \\ \midrule
		v2t\_navigator~\cite{Jin:2016hz}     & 42.6  & 28.8   & 61.7     & 46.7  \\
		Aalto~\cite{Shetty2016}    & 39.8   & 26.9  & 59.8    & 45.7 \\ 		
		VideoLAB~\cite{Ramanishka2016}                               & 39.1   & 27.7  & 60.6    & 44.1 \\ 
		ruc-uva~\cite{dong2016early}                               & 38.7   & 26.9  & 58.7    & 45.9 \\
		Dense Caption~\cite{shen2017weakly}      & 41.4  & 28.3   & 61.1     & 48.9  \\ \hline
		Our XE                   & 43.0   & 28.7  & 61.7    & 47.3 \\
		CST\_GT\_None                 & \textbf{44.1}   & \textbf{29.1}  & \textbf{62.4}    & \textbf{49.7} \\ \midrule
		\multicolumn{5}{c}{Full RL training} \\ \midrule
		CIDEnt-RL~\cite{pasunuru2017reinforced}  & 40.5  & 28.4   & 61.4     & 51.7  \\ 
		SCST & 41.3   & 28.1  & 61.9    & 51.9 \\
		\hline
		CST\_MS\_Greedy & \textbf{42.2}   & \textbf{28.9}  & \textbf{62.3}    & 53.4 \\
		CST\_MS\_SCB    & 41.9   & 28.7  & 62.1    & 53.6 \\
		CST\_MS\_SCB (*)   & 41.4   & 28.8  & 62.2    & \textbf{54.2} \\
		\bottomrule
	\end{tabular}
\end{table}

\begin{table*}[th]
	\centering
	\caption{Performance of CST\_GT\_None when optimizing different metrics (different reward functions).}
	\label{cst_others}
	\begin{tabular}{@{}Hl|ccccccc@{}}
		\toprule
		Run                   & Optimization metric   & BLEU\_1 & BLEU\_2 & BLEU\_3 & BLEU\_4 & METEOR & ROUGE\_L & CIDEr \\
		\midrule
		resnetc3dmfcccategory &   XE     &  82.1   & 68.6   & 55.2   & 43.0   & 28.7  & 61.7    & 47.3 \\
		\midrule
		resnetc3dmfcccategory & BLEU\_4  & 80.6   & 67.1   & 53.9   & 41.9   & 28.4  & 61.1    & 45.5 \\
		resnetc3dmfcccategory & METEOR   & 81.6   & 68.1   & 54.8   & 42.6   & 28.7  & 61.4    & 47.7 \\ 		
		resnetc3dmfcccategory & ROUGE\_L & 82.3   & 69.1   & 55.8   & 43.9   & 29.0  & 62.1    & 48.6 \\
		resnetc3dmfcccategory &   CIDEr     &\textbf{82.8}   & \textbf{69.5}   & \textbf{56.2}   & \textbf{44.1}   & \textbf{29.1}  & \textbf{62.4}    & \textbf{49.7} \\
		\bottomrule
	\end{tabular}
\end{table*}

\subsection{Comparison Against the State of the Art}
Table~\ref{tab:sota} reviews the leading results from previous literature on the MSRVTT video captioning benchmark in terms of four different scoring metrics (BLEU~\cite{papineni2002bleu}, METEOR~\cite{lavie2014meteor}, ROUGE\_L~\cite{lin2004rouge}, and CIDEr~\cite{vedantam2015cider}), and compares this to the performance of XE and CIDEr-optimizing models trained as described in Sec.~\ref{ssec:systems}.
On three of the four metrics, v2t\_navigator~\cite{Jin:2016hz} -- a system based on cross-entropy minimization -- achieves the highest score. On the CIDEr metric, our primary interest since it correlates best with human judgments, the situation is different, however: the best score is reported by CIDEnt-RL~\cite{pasunuru2017reinforced}, the only video-captioning publication we are aware of that optimizes CIDEr score directly, through RL. Their score, 51.7, is the highest CIDEr score reported on this benchmark thus far.

Among the methods implemented for this paper, our conventional XE system performs very similarly (and often slightly better than) v2t\_navigator on all metrics. The CST-derived methods improve further compared to this baseline. In particular, our proposed pre-training CST\_GT\_None (a.k.a.\ WXE) outperforms all other methods on the metrics that are not CIDEr, while also improving the CIDEr score by more than 2 points. Nonetheless, we are most interested in the CIDEr metric, since it better captures human consensus. We see that fine-tuning CST\_GT\_None using the two REINFORCE-schemes implemented leads to further improved CIDEr scores. Starting full RL training from WXE rather than XE (CST\_GT\_Greedy vs.\ SCST) actually improved performance on all measures, establishing the utility of the WXE pre-training scheme.
The highest CIDEr score was attained by our self-consensus baseline. This is appealing, since SCB represents a simple idea and runs about twice as fast as CST\_MS\_Greedy. Our full CST proposal gives a CIDEr score of 53.6, almost 4 points greater than WXE, and 2 points superior to the best score from prior literature. In the last run, denoted by (*), we experimented with computing the SCB from the 20 sequences sampled during training, instead of the 20 GT sequences.\footnote{The gradient estimate in~\eqref{eq:addbaseline} may now be biased, but the computation time is unchanged as we anyhow need to compute CIDEr for the samples.} This reached a new state-of-the-art CIDEr score of 54.2.

\subsection{CST for Other Rewards}
Even though CST was conceived for optimizing metrics that reward consensus (like CIDEr), we also investigated the efficacy of the CST\_GT\_None training scheme in optimizing reward functions based on other objective scoring metrics. As shown in Table~\ref{cst_others}, CIDEr and ROUGE\_L are the only two reward functions where CST outperforms XE under all evaluation metrics. Among those two, CIDEr optimization scores the best.


\begin{figure}[t]
	\begin{center}
		\includegraphics[width=1\linewidth]{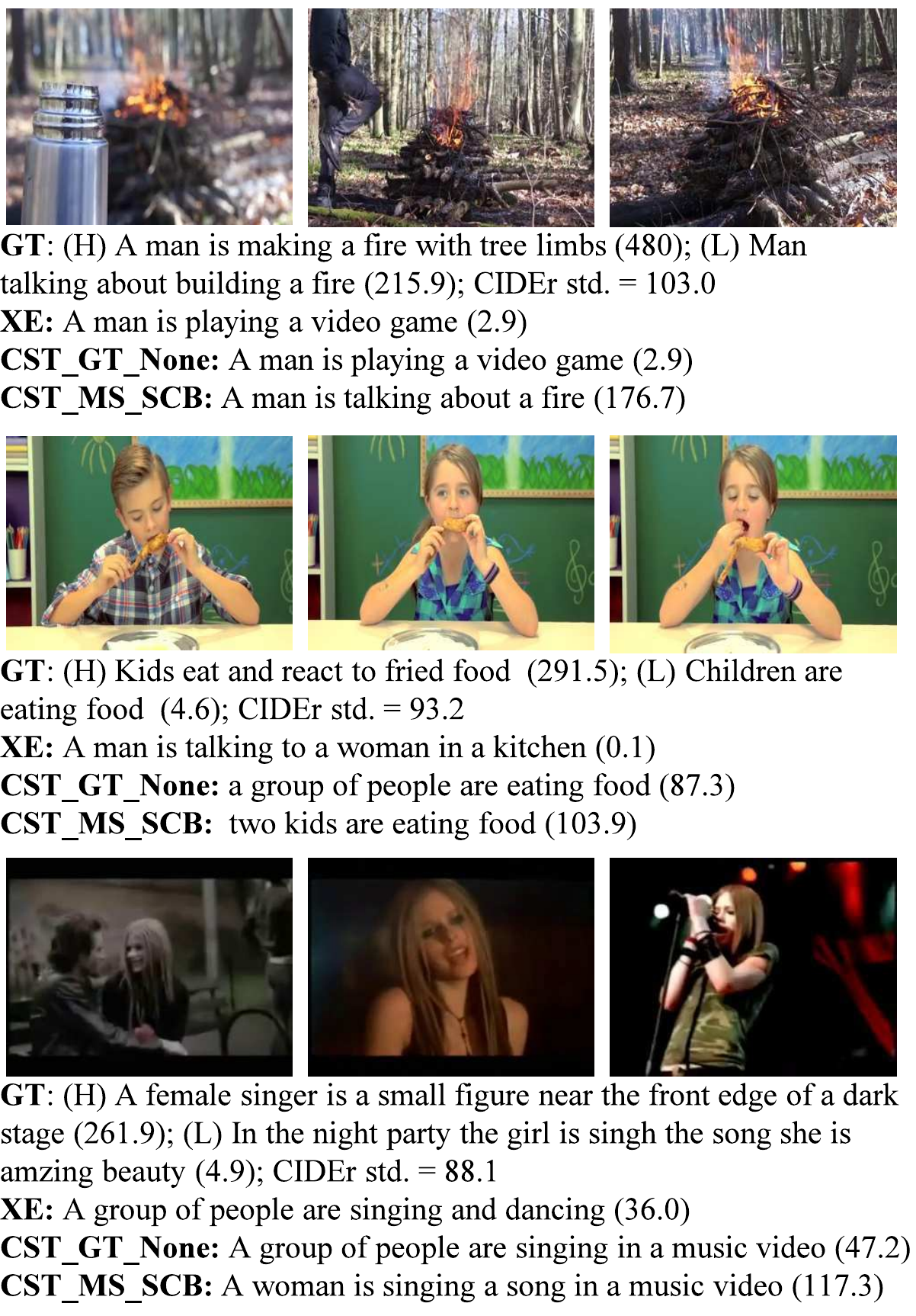}
	\end{center}
	\caption{Example captions for videos with a broad range of CIDEr scores (from L to H) among human annotators.}
	\label{fig_examples}
\end{figure} 

\subsection{Qualitative Analysis of Generated Captions}
Figure~\ref{fig_examples} shows examples of generated captions and their CIDEr scores for videos where human reference annotations had a high standard deviation in held-out CIDEr score.
These are videos where some human captions are close to a consensus, while others deviate considerably from it.
On this type of video, the XE training is likely to be unduly influenced by outlying examples, leading to poor output, as seen in the figure.
In contrast, CST, especially with full RL, tends to generate more appropriate captions.

\section{Conclusion}
\label{sect:conclusion}
This paper proposed a Consensus-based Sequence Training (CST) scheme to generate video captions. CST is a variant of the REINFORCE algorithm, but exploits the multiple existing training-set captions in several new ways.
First, based on a new theoretical insight, CST performs an RL-like pre-training, but with captions from the training data replacing model samples. This alleviates the objective mismatch issue in conventional XE training while being equally fast.
Second, CST applies REINFORCE for fine-tuning using the consensus (average reward) among training captions as the baseline $b$. This fine-tuning additionally removes the exposure bias, and trains twice as fast as a greedy baseline estimator. The two stages of CST allow objective mismatch and exposure bias to be assessed separately, and together establish a new state-of-the-art on the task.

{\small
\bibliographystyle{ieee}
\bibliography{egbib,rl}
}

\newpage


\appendix{\textbf{\large Supplemental Material}} 
\begin{table*}[h!]
	\centering
	\caption{Performance comparison of our WXE (CST\_GT\_None) model and standard XE model on different features.}
	\label{table_cst}
	\begin{tabular}{@{}l|c|ccccccc@{}}
		\toprule
		Input feature(s)                   & Method & BLEU\_1 & BLEU\_2 & BLEU\_3 & BLEU\_4 & METEOR & ROUGE\_L & CIDEr \\ \midrule
		\multirow{2}{*}{Resnet}                &  XE  & 75.6   & 61.0   & 48.1   & 37.2   & 26.7  & 58.0    & 39.9 \\		
		&  CST\_GT\_None      & 75.8   & 61.3   & 48.5   & 37.5   & 26.6  & 58.3    & 41.5 \\
		\midrule		
		\multirow{2}{*}{C3D}                   &   XE     & 77.2   & 62.4   & 48.9   & 37.2   & 26.7  & 58.5    & 39.8 \\
		&   CST\_GT\_None     & 77.5   & 62.9   & 49.6   & 38.1   & 27.1  & 59.0    & 43.4 \\		
		\midrule		
		\multirow{2}{*}{MFCC}                  &    XE    & 68.9   & 49.8   & 36.4   & 26.1   & 21.8  & 51.3    & 17.9 \\		
		&    CST\_GT\_None    & 69.5   & 51.1   & 38.1   & 27.7   & 22.0  & 51.8    & 19.8 \\
		\midrule		
		\multirow{2}{*}{Category}              &    XE    & 70.2   & 52.1   & 38.8   & 28.4   & 22.7  & 53.0    & 25.5 \\
		&    CST\_GT\_None    & 69.1   & 50.1   & 37.2   & 27.2   & 22.6  & 52.2    & 24.2 \\		
		\midrule		
		\multirow{2}{*}{\begin{tabular}[c]{@{}l@{}} ResNet + C3D \\ + MFCC + Category\end{tabular}} &   XE     & 82.1   & 68.6   & 55.2   & 43.0   & 28.7  & 61.7    & 47.3 \\
		&   CST\_GT\_None    & \textbf{82.8}   & \textbf{69.5}   & \textbf{56.2}   & \textbf{44.1}   & \textbf{29.1}  & \textbf{62.4}    & \textbf{49.7} \\		
		\bottomrule
	\end{tabular}
\end{table*}

\section{Further Training Details of CST}
We encountered the same problem reported in ~\cite{Rennie2017CVPR} when training our CST models. More specifically, the generated captions tends to end unexpectedly such as ``a girl is applying makeup to her'', ``a woman is talking to a'', and ``a woman is talking about a". We suspect the models get difficult in predicting the correct ending token, while those bigrams such as ``to her", ``to a" and ``about a" were common among the captions, thus influences the CIDEr rewards. As suggested by~~\cite{Rennie2017CVPR}, we also include $<$EOS$>$ at the end of each caption in the reference set when computing CIDEr rewards. This simple amendment substantially prevents the model from generating those incomplete sentences. 

\section{Derivation of Eq.~\eqref{eq:dxe}}
In~\cite{Ranzato2016ICLR} the authors did not provide full derivations of Eq.~\eqref{eq:dxe} and Eq.~\eqref{eq:drl}. For the sake of completeness, in this section and the next section we provide full derivations for both equations.

Because $o_t$ is softmax input (the logit), we can compute the probability distribution over the vocabulary at time step $t$ as follows:
\begin{equation}
	p_\theta (w_t | h_t) = \frac{\exp(o_t)}{\sum \exp(o_t)}
\end{equation}
In Eq~\eqref{eq:xeloss}, to compute the XE loss, we also use $p_\theta (w_t | h_t)$ in the context of the probability of the ground-truth word $w_t$. Recall that $1(w_t)$ is the one-hot representation of $w_t$, the XE loss can be rewritten w.r.t. $o_t$ as follows:

\begin{align}
	L^{XE}(\theta) &= - \sum_{t=1}^{T} \log \frac{\exp(o_t^\intercal 1(w_t))}{\sum \exp(o_t)},
\end{align}

Note that $\frac{\partial L^{XE}(\theta)}{\partial o_t}$ depends only on the softmax input $o_t$. Therefore:

\begin{align}
	\frac{\partial L^{XE}(\theta)}{\partial o_t} &= - \frac{\partial }{\partial o_t} \bigg( \log \frac{\exp(o_t^\intercal 1(w_t))}{\sum \exp(o_t)} \bigg) \nonumber \\
	&= \frac{\partial \log {\sum \exp(o_t)}}{\partial o_t}  - \frac{\partial o_t^\intercal  1(w_t)}{\partial o_t} \nonumber \\
	&= \frac{1}{\sum \exp(o_t)}  \frac{\partial \sum \exp(o_t)}{\partial o_t}   - 1(w_t) \nonumber \\
	&= \frac{\exp(o_t)}{\sum \exp(o_t)} - 1(w_t) \nonumber \\
	&= p_\theta (w_t | h_t) - 1(w_t)
	\label{eq:derivation_xe}
\end{align}


\section{Derivation of Eq.~\eqref{eq:drl}}
From Eq~\eqref{eq:addbaseline}, we can compute $\frac{\partial L^{RL}(\theta)}{\partial \theta}$ by a single sample sequence $w^s$:
\begin{align}
	\frac{\partial L^{RL}(\theta)}{\partial \theta} &= - (r(w^s) - b) \frac{\partial \log p_\theta(w^s)}{\partial \theta} \nonumber \\
	&= - (r(w^s) - b) \sum_{t=1}^{T} \frac{\partial \log p_\theta(w^s_t | h_t)}{\partial \theta} 
\end{align}

Here we also use $p_\theta (w^s_t | h_t)$ in the context of the probability of the sampled word $w^s_t$. More precisely, $\frac{\partial L^{RL}(\theta)}{\partial \theta}$ can be computed in term of $o_t$ and the one-hot representation $1(w^s_t)$ as follows. 

\begin{align}
	\frac{\partial L^{RL}(\theta)}{\partial \theta} &= - (r(w^s) - b) \sum_{t=1}^{T}  \frac{\partial }{\partial o_t} \bigg( \log \frac{\exp(o_t^\intercal 1(w^s_t))}{\sum \exp(o_t)} \bigg)
\end{align}

Note that the reward function $r(.)$ and the baseline $b$ are treated as constants during back-propagation, and $\frac{\partial L^{RL}(\theta)}{\partial o_t}$ depends only on the logit $o_t$. Therefore:

\begin{align}
	\frac{\partial L^{RL}(\theta)}{\partial o_t} &= - (r(w^s) - b) \frac{\partial }{\partial o_t} \bigg( \log \frac{\exp(o_t^\intercal 1(w^s_t))}{\sum \exp(o_t)} \bigg)
\end{align}
Following the same steps in~\eqref{eq:derivation_xe} to take the gradient of the second term, we come up with Eq.~\eqref{eq:drl}.

\section{Performance on the Validation Set}
\begin{figure}[t!]
	\begin{center}
		\includegraphics[width=1\linewidth]{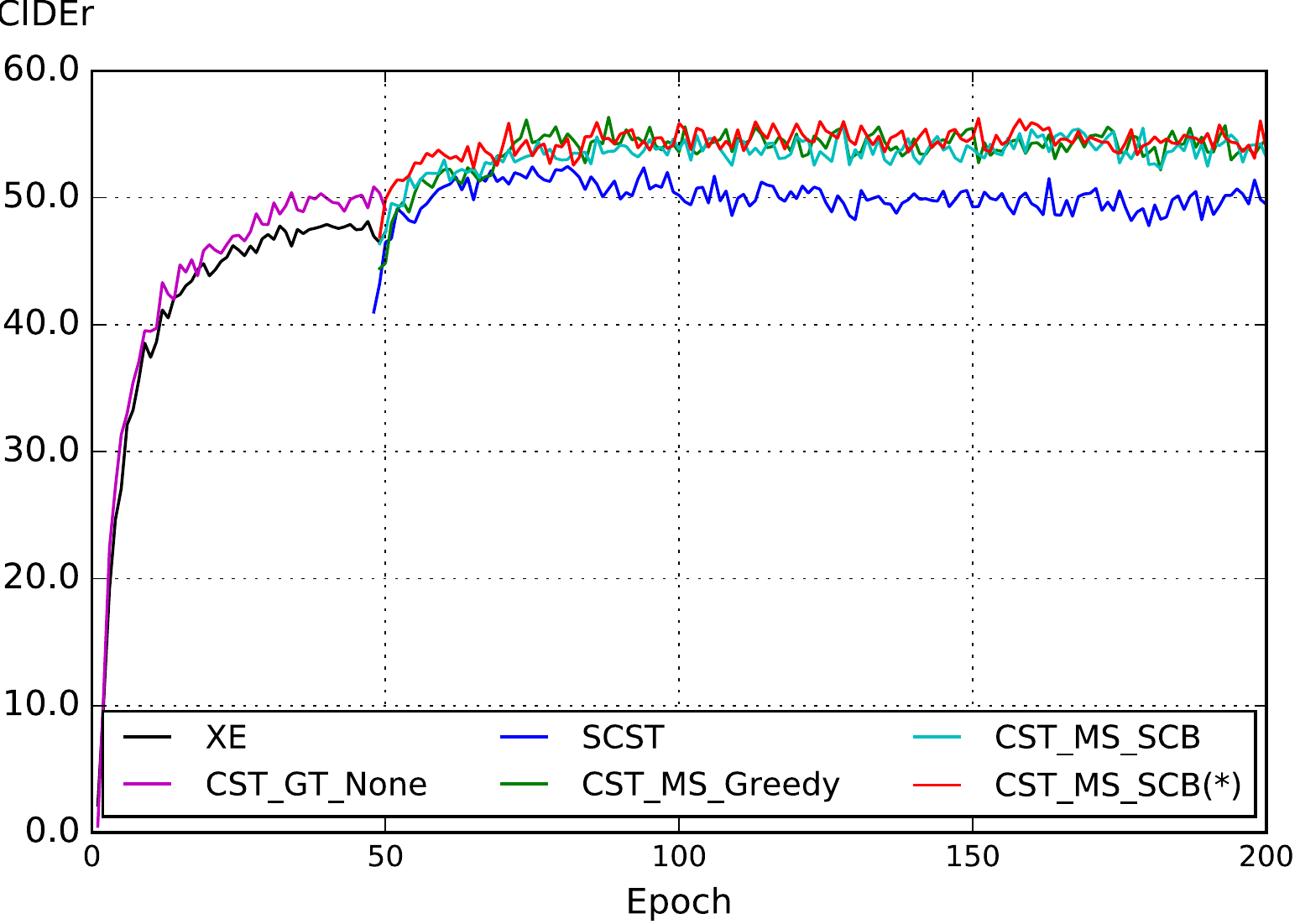}
	\end{center}
	\caption{Performance on the validation set during training of different methods.}
	\label{fig:validation}
\end{figure} 
We check the CIDEr performance on the validation set for every epoch and report the performance in Fig~\ref{fig:validation}. Our WXE model (CST\_GT\_None) almost always reaches better check points than the standard XE model (in the first 50 epochs). When switching from XE training to RL training, all the RL models get difficult to converge in the first few iterations, but quickly recover and reach higher optimal levels. When start RL training from our WXE model, both the SCB and greedy baselines have similar convergence rates; however, our SCB baselines are two times faster to compute. Performance of  SCST~\cite{Rennie2017CVPR} model in our implementation, which was started training from the XE model, is clearly inferior to the RL models which was started from the pre-train WXE model.

\section{WXE vs. XE Training}
In Table~\ref{table_cst} we compare the performance of our WXE (CST\_GT\_None) with the standard XE training on different video features. As we see, WXE significantly improves CIDEr metrics for all feature except Category feature. For example, the performance gain on Resnet, C3D and MFCC features are 1.6, 3.6 and 1.9 CIDEr respectively. We also observe that optimizing for CIDEr metric will also improve other metrics in most of the cases. Performance of Category feature is not reliable to assess the effectiveness of our WXE method because there are only 20 different categories, and many videos have the same category label. Therefore, this feature is not discriminative and should not be used alone to evaluate the captioning model. In summary, our proposed WXE method performs better than the standard XE method. Especially, once the rewards such as CIDEr scores are pre-computed, the WXE loss is as fast to compute as the standard XE loss.

%
%


\end{document}